\documentclass[conference]{ieeeconf}

\IEEEoverridecommandlockouts                              
\usepackage{todonotes}
\newcommand{\jiawei}[1]{{\color{orange}#1}}

\newcommand{\jeff}[1]{\todo[inline, backgroundcolor=orange!60]{Jeff: #1}}

\usepackage[T1]{fontenc}
\usepackage{amsfonts}
\usepackage{amsthm}
\usepackage{pifont}
\usepackage{xcolor}
\usepackage{soul}

\newtheorem{definition}{Definition}

\newtheorem{problem}{Problem}
\usepackage{glossaries}
\usepackage{listings}
\usepackage{amsmath}
\usepackage{amssymb}
\usepackage{subfig}
\usepackage{graphicx}
\usepackage{stix}
\usepackage{tikz}
\usepackage{etoolbox}
\usetikzlibrary{shapes,arrows}
\usepackage{verbatim}
\usepackage[hidelinks]{hyperref}

\usepackage{soul}
\newcommand{\david}[2]{{\st{#1}} {\color{gray} #2}}

\title{\LARGE \bf
Toward Fine Contact Interactions: Learning to Control Normal Contact Force with Limited Information
}

\author{Jinda Cui$^{\dag}$, Jiawei Xu$^{\ddag}$, David Saldana$^{\ddag}$, and Jeff Trinkle$^{\ddag}$
\thanks{*This work was partially supported by the National Science Foundation through EFRI C3 SoRo (award 1832795)}
\thanks{$^{\ddag}$Autonomous and Intelligent Robotics Laboratory (AIRLab) and the Department of Computer Science and Engineering, Lehigh University, Bethleham, PA, USA. \url{<jix519, das819, jct519>@lehigh.edu}}%
\thanks{$^{\dag}$Honda Research Institute USA, San Jose, CA, USA. (Work was done during the Ph.D. study at Lehigh AIRLab). \url{jinda.cui@gmail.com}}
}

\begin{document}

\maketitle
\thispagestyle{empty}
\pagestyle{empty}

\begin{abstract}
Dexterous manipulation of objects through fine control of physical contacts is essential for many important tasks of daily living. A fundamental ability underlying fine contact control is compliant control, \textit{i.e.}, controlling the contact forces while moving. For robots, the most widely explored approaches heavily depend on models of manipulated objects and expensive sensors to gather contact location and force information needed for real-time control. The models are difficult to obtain, and the sensors are costly, hindering personal robots' adoption in our homes and businesses. This study performs model-free reinforcement learning of a normal contact force controller on a robotic manipulation system built with a low-cost, information-poor tactile sensor. Despite the limited sensing capability, our force controller can be combined with a motion controller to enable fine contact interactions during object manipulation. Promising results are demonstrated in non-prehensile, dexterous manipulation experiments. 

\end{abstract}

\section{Introduction}

Deep learning has advanced robot manipulation in unstructured human environments dramatically over the past few years \cite{kroemer2021areview}. The adaptability of learning robots in the face of ubiquitous variations is a major step toward capable personal robots~\cite{cui2021toward}. Most research in robot learning for manipulation~\cite{kroemer2021areview} focuses on stable static grasps that maintain fixed contacts with the object. However, learning robots are still in their infancy when it comes to performing manipulation tasks requiring contact interactions including transitions between sticking, slipping, and rolling \cite{balkcom2002wrench}. Such contact interactions may have occurred in dexterous manipulation tasks \cite{openai2020learning}, but the robots are clumsy in comparison to humans. In fact, one can consider the learned dexterous manipulation skills in~\cite{openai2020learning} as active caging \cite{bircher2021complex}. In other words, the cage moves and changes shape over time to force the object to move to its goal configuration.  By contrast, a robot able to control contact interactions could leverage those skills to achieve the same goal in ways similar to what a human would do.


We call the act of controlling contact interactions \emph{fine contact interactions}.  Achieving and breaking contacts intentionally and choosing to maintain sticking, slipping, or rolling contact at one or more contacts and transitioning among them are examples of fine contact interaction. Robust execution of fine contact interactions in uncertain environments requires the simultaneous control of position and force \cite{lozano-perez1984automatic}, such as combined force and position control \cite{lefebvre2005active} and combined force and motion control \cite{hou2019robust}. 

\begin{figure}[t]
    \centering
    \includegraphics[width=0.4\textwidth]{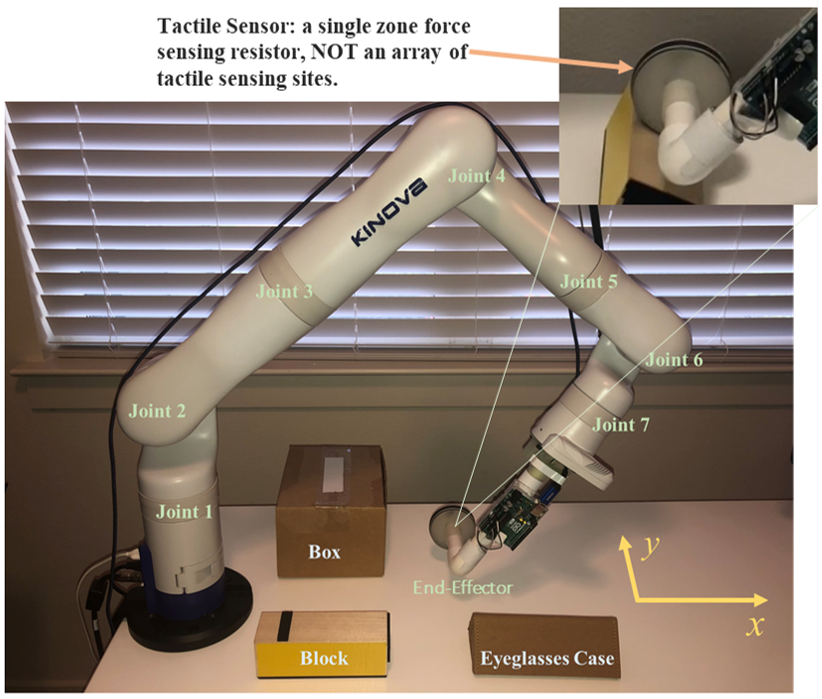}
    \caption{Environment for hardware experiments: A Kinova GEN3 robot, a low-cost tactile sensor, and different types of objects. }
    \vspace{-1em}
    \label{fig:robot_GEN3}
\end{figure}

When the force control regulates the normal contact force, and the motion control focuses on the movements tangential to the contact, the robot can actively control and change the transition of contact interactions - realizing fine contact interactions. Such a controller is essentially a hybrid force and velocity controller \cite{hou2019robust}, which requires the identification of orthogonal force and motion directions. 
The identification of these directions usually requires precise models and advanced contact sensing capabilities. For example, Hou and Mason enabled contact-maintaining non-prehensile manipulation by solving an optimization problem at each time-step \cite{hou2019robust}. Object models were used to produce force- and motion-controlled directions hence the robot actions. Hogan et al., also using optimization, developed contact mode-constrained ``manipulation primitives'' such as object sliding, pushing, and tilting. Their method relies on real-time availability of object and contact models \cite{hogan2020tactile}. In addition, the sensor requirements are high: Hou and Mason used a force-torque sensor combined with a customized end-effector and the measurement is tied to the model of the end-effector \cite{hou2019robust}.  Hogan et al. used an optical-based tactile sensor which is not readily available to the general public \cite{hogan2020tactile}. The BioTac sensor used by Sutanto et al. is expensive and delicate  \cite{sutanto2019learning}.
The limitation on model availability and sensing capacity prevent model-based analytical methods from performing well in unstructured environments on low-cost personal robots.






In this paper, we present a model-free deep reinforcement learning (DRL) approach to learn a normal contact force controller which can be combined with an independent motion controller (similar to parallel force and position control \cite{chiaverini1993parallel}) to perform fine contact interactions for a single contact point. The novelty of our research is that we focus on force control on robots equipped with low-cost, information-poor sensors with limited model assets which cannot be addressed using traditional analytical approaches. An example system is shown in Fig.~\ref{fig:robot_GEN3}. The robot is a collaborative robot popular among research labs, and the low-cost  sensor on the end-effector only measures the magnitude of the normal contact force. We emphasize that despite the manipulator and the joint used in training, the learned force control skill may generalize to a different joint or a different manipulator without retraining due to the nature of model-free learning.

Our contributions are: 1) We provide a novel formulation of the control policy;
2) We propose and implement an effective method to stabilize the training performance; and
3) We perform training in a simple virtual environment, and show that the trained policy can directly deploy on a practical robot arm, demonstrating force control and fine contact interactions in the real world. 

\section{Problem Statement}



\noindent

\begin{figure}[t]
    \centering
    \vspace{0.5em}
    \includegraphics[width=0.42\textwidth]{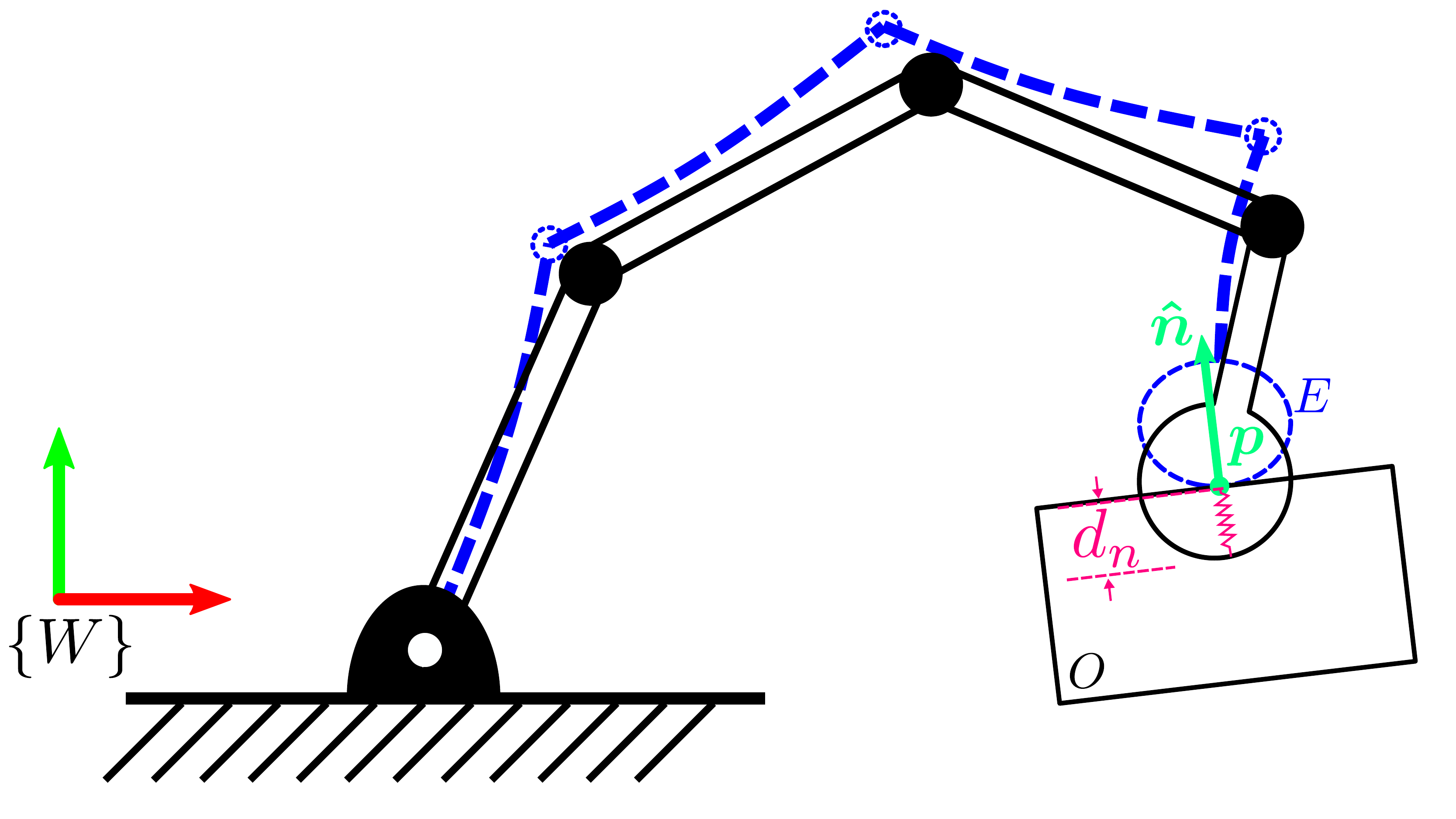}
    \caption{
    The end-effector $E$ contacts an object $O$ at point $\mathbf{p}$ generating normal force $f_n$ in contact normal direction $\mathbf{\hat n}$. If the robot joints were rigid and the object were not present, its configuration would be that shown in black. The blue dashed configuration shows (and exagerates) the fact that the robot deflects due to the contact force. 
    } 
    \label{fig:lumpedSpring}
    \vspace{-1.5em}
\end{figure}

We use an $n$-DOF robotic arm as the \emph{manipulator} which has multiple links connected by revolute joints. Considering a world fixed frame denoted by $\{W\}$, the first link of the manipulator is rigidly attached to a base that is static in $\{W\}$, and the last link is attached to an end-effector $E$ to manipulate objects. The joint position $\mathbf{q}\in\mathbb{R}^n$ determines the arm configuration, which determines the pose of~$E$. The robot takes joint velocity commands $\mathbf{\dot{q}}$ as input. Fig.~\ref{fig:lumpedSpring} illustrates a planar robot manipulating an object $O$ through contact with its circular end-effector~$E$.  

Suppose a combined force/motion controller sends a motion command $\mathbf{\dot{q}}^d$ to the robot to execute a desired end-effector motion. The normal force controller's task is to alter $\mathbf{\dot{q}}^d$ in the normal contact direction $\mathbf{\hat{n}}$ to maintain the desired normal contact force $f_{n}^{d}$.  It does this by adjusting the nominal velocity  as follows: $\mathbf{\dot{q}'} = \mathbf{\dot{q}}^d + \mathbf{\dot{q}}^{f\!c}$ (see Fig.\ref{fig:system_diagram}). 

It is challenging to generate $\mathbf{\dot{q}}^{f\!c}$ analytically. The relationship between $\mathbf{\dot{q}}^{f\!c}$ and the normal force response depends on five critical pieces of information that may not be available to the low-cost personal robots. First, the magnitude of the normal contact force $f_{n}$. Second, the position of the contact point $\mathbf{p}$. Third, the contact normal direction $\mathbf{\hat{n}}$. Fourth, the system compliance, represented as the lumped spring in Fig.~\ref{fig:lumpedSpring}. The lumped spring summarizes the robot compliance and the contact compliance at $\mathbf{p}$ in the normal direction with a spring deformed by $d_{n}$, while assuming everything else matches the rigid model~\cite{Prattichizzo2016}. Fifth, the motion at the contact in the normal direction caused by the motion controller or the object motion resulted from external forces.

In this paper, we study normal force control for scenarios under which the environment is unknown and the manipulator's sensors are unable to provide the necessary information to implement force control analytically.
Instead of using an expensive tactile sensor, we propose to use a low-cost tactile sensor that covers the entire end-effector, which is only able to measure the magnitude of the contact force, but not its position or its normal direction.

\begin{problem}[\textbf{Marginally-informed Normal Force Control}]
    Find the control output $\mathbf{\dot{q}}^{f\!c}$ to control $f_n$ to track a desired normal force $f_n^d$ during object manipulation under the following assumptions.
    \textit{i)} There is a single contact point between the sensor and the object.
    \textit{ii)} The manipulator is executing an arbitrary velocity command $\mathbf{\dot{q}}^{d}$.
    \textit{iii)} The magnitude of the normal force $f_{n}$ is \textbf{observable}.
    \textit{iv)} The direction of the normal force and the location of the contact point are \textbf{unknown}.
    \textit{v)} The models of the object, the environment, and the robot are \textbf{unknown}.

    \label{p:1}
\end{problem}

Doing force control with all joints of the manipulator would require the DRL agent to explore the robot's $n$-dimensional configuration space, which would significantly increase the search space of the learning problem. Therefore, in this paper, we focus on a \textbf{Reduced Problem 1} - achieving the marginally-informed normal force control by adjusting the velocity output of a single joint.

\begin{figure}[t]
    \centering
    \vspace{.8em}
    \includegraphics[width=0.4\textwidth]{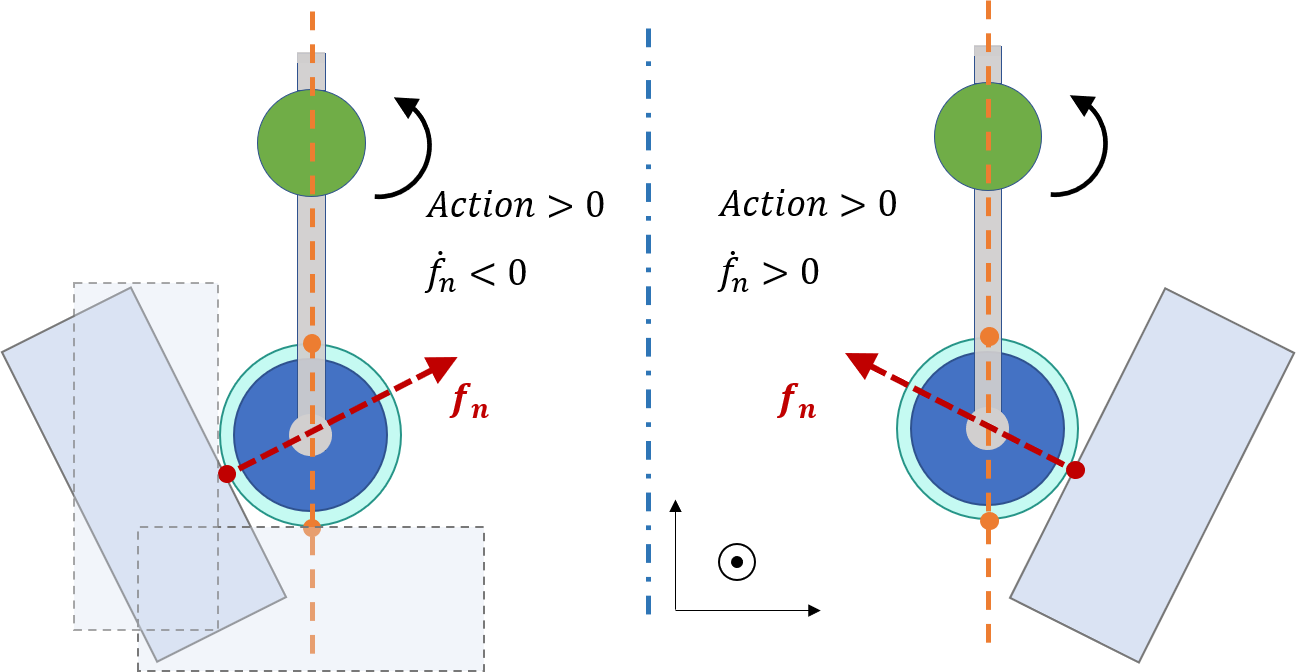}
    \caption{The effect of joint velocity on the normal force is strongly affected by contact location and normal direction.}
    \label{fig:chiral_1}
    \vspace{-0.8em}
\end{figure}

Learning this reduced force control problem is still challenging. Fig.~\ref{fig:chiral_1} illustrates two scenarios where a box is in contact with the sensor (light blue) at various locations. The joint shown in green is controlled by the control policy to regulate the normal contact force. 

On the left illustration, we can see that the relationship between the joint action and the normal force (the red arrow) changes as the contact point on the sensor changes. If the contact point is on the right side, as shown on the right illustration, the action-force relationship flips the sign. The two points on the end-effector intersected by the orange dashed line are affected by the boundary singularity on which the force control ability is lost. Despite this limitation, the force control policy must adapt to the variations in contact location and, therefore, the variations in the action-force relationship, even if contact location and normal force direction are not measured, the models are unknown, and the motion controller may disturb the force control. Nonetheless, the action-force relationship may be inferred through the interactions between the robot and the object. Therefore, we take advantage of the robot's interaction capabilities and tackle this problem using model-free deep reinforcement learning.

\section{{Reinforcement} Learning Approach}
\noindent


\begin{figure*}[t]
    \centering
    \includegraphics[width=0.92\textwidth]{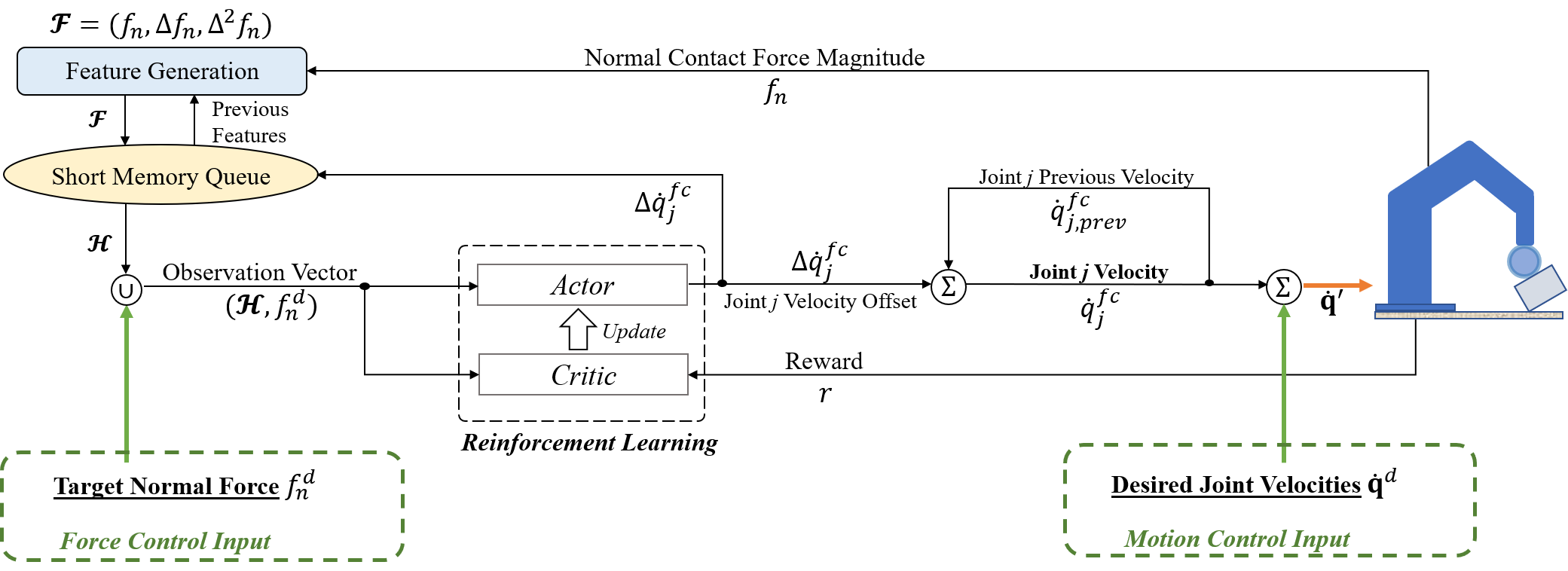}
    \caption{System Diagram: the learned force controller's task is to track the target normal force during object manipulation. When desired joint velocities present, the force controller adjusts the velocity at one robot joint, which modifies the motion at the contact to achieve the target normal force.} 
    \label{fig:system_diagram}
    \vspace{-1em}
\end{figure*}

\noindent
We aim at a sample-efficient learning without complicated reward engineering. Therefore, we base our approach on a mainstream continuous action space DRL method - Deep Deterministic Policy Gradient (DDPG) \cite{lillicrap2019continuous}. DDPG is an off-policy learning method known to be more sample-efficient than on-policy methods such as PPO \cite{schulman2017ppo}. We use Hindsight Experience Replay (HER) with DDPG to further enhance the sample efficiency, which has proven track records in learning robot manipulation tasks \cite{marcin2017hindsight}. On top of DDPG and HER, we propose the following structure and constraints to learn the Marginally-informed Normal Force Controller:
\begin{itemize}
    \item Interaction-conditioned policy observation, which allows information-probing actions to emerge in training.
    
    \item Joint velocity offset instead of joint velocity as the policy's action to simplify the learning problem.
    
    \item A simple virtual training environment with effective domain randomization designs to avoid simulating the more complex real-world robot.
    
    \item Task-Balanced Network Update to stabilize the training process and improve learning performance.
\end{itemize}

\noindent
The overall system diagram is shown in Fig.~\ref{fig:system_diagram}. The actor and critic blocks are the neural networks being trained, where the actor is our control policy, and the critic is used to update the actor. The force controller takes the target normal force and measurement features as input, and its output is summed with motion commands (from a motion controller) for the robot to execute. We explain the details of the policy next.




\subsection{Observation and Action}
\noindent
Generally speaking, a controller may adapt to changes that are not directly observed by inferring information from the responses of its control actions. However, when a system is severely limited in observability (such as our system), the control actions may not gather enough information. Instead, as pointed out in \cite{feldbaum1963dual}, a controller may execute ``probing actions'' that do not directly drive the system to the control target, but instead, probe for more information. We aim to enable our marginally-informed force controller to gather information actively through probing actions while seeking to achieve the desired normal force. 

A prerequisite for probing actions to emerge during reinforcement learning is that the agent can observe the interactions. Therefore, we let the actor and critic observe the most recent force interaction history of length $N$, denoted by~$\mathcal{H} = \{(a_{c-i},\boldsymbol{\mathcal{F}}_{c-i})\vert i\in[1,N]\}$, where $a$ is the action, $\boldsymbol{\mathcal{F}}$ is a feature vector of the force response, and $c$ is the index of the current time-step.  The feature vector is $\boldsymbol{\mathcal{F}}_{c-i} = (f_{n},\Delta f_{n},\Delta^2 f_{n})_{c-i}$. The tactile sensor measures the normal force magnitude $f_{n}$ at each time-step. The discrete-time derivatives of $f_{n}$, $\Delta f_{n}$ and $\Delta^2 f_{n}$ can provide information about the changes at the contact. The desired normal contact force $f^d_{n}$ is observable for the policy. Thus, the full observation as the input to the policy network is $\mathcal{O}=(\mathcal{H},f^d_{n})$. 

We set the policy's action to be $a = \Delta\dot{q}_{j}^{f\!c}$, which is a joint velocity offset at joint $j$. The velocity offset is added to the previous joint velocity command before being combined with the motion command. We made this design choice to simplify the learning problem: first, we combine $\Delta\dot{q}_{j}^{f\!c}$ with $\dot{q}_{j,prev}^{f\!c}$ instead of the previous $\dot{q}'_{j}$ to decouple the force and motion controllers since $\dot{q}'_{j}$ is partially contributed by $\dot{q}^{d}_{j}$. 
Second, by making the policy's direct action a joint velocity offset instead of a joint velocity, we avoid a more complex learning formulation conditioned on joint velocity: the normal contact force control can be viewed as deformation control facilitated by relative motions between the robot and the object at the contact (recall the lumped spring example). Such relative motions are not fully determined by the joint velocity due to other sources of motion, such as the object and the motion controller. Since the policy's past actions appear in its observation, using joint velocity as the action adds distracting information to the observation instead of revealing the status of the relative motion.

\subsection{Training Environment and Domain Randomization}

\noindent
The success of deep reinforcement learning for robot manipulation has been achieved largely through training in simulation. In general, training in simulation offers safety, speed (time-stepping can be faster than real-time), and flexibility (easier to set up and modify). Since learning force control is potentially dangerous to the hardware, training in simulation is a natural choice. In fact, we can easily set up a simple environment in simulation without the need to simulate the actual robot to learn the force control skill.  

\begin{figure}[t]
    \centering
    \vspace{.2em}
    \includegraphics[width=0.45\textwidth]{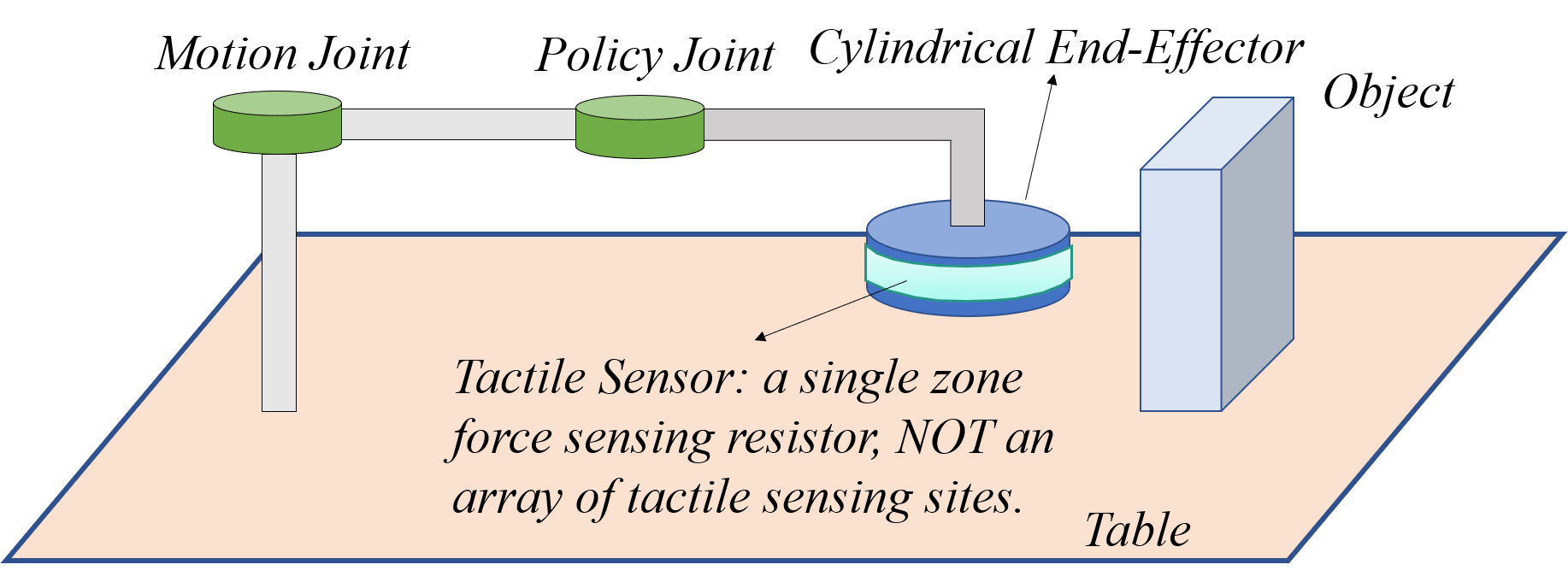}
    \caption{Illustration of the training environment.}
    \label{fig:robot}
    \vspace{-1em}
\end{figure}

The simulated environment is illustrated in Fig.~\ref{fig:robot}. The robot has two velocity-controlled revolute joints and one cylindrical end-effector. The virtual end-effector and the sensor match the hardware in dimension and functionality. The force control policy controls one robot joint, referred to as the ``policy joint,'' to regulate the normal force, and the other joint, referred to as the ``motion joint,'' to simulate the disturbances from a motion controller. The policy joint's velocity-force relationship is approximately matched to the hardware robot (with joint 6 being the policy joint) for a single time-step. 

During training, the policy interacts with the environment during policy roll-outs {(\textit{i.e.}, simulations of $T$ time-steps)}. At each time-step, the agent makes observation and computes an action for the ``policy joint'' to execute. A reward is given at each time-step. We use the two simple rewards in training: the negated absolute difference between measured and target forces: $r_{f} = -\vert f_{n} - f^d_{n}\vert$, and a break contact penalty: $r_{c} = -M$, where $M\in\mathbb{R}_+$ is a large positive number that emphasizes on the importance of maintaining the contact. The latter reward replaces the prior when the contact is broken.

Environment reset is performed at the beginning of each policy roll-out, during which initial contact is made between the robot and the box's surface. We perform domain randomization (DR) during the reset. The parameters being randomized (within predefined ranges) are the initial contact normal angle (thus the contact location on the sensor), initial normal contact force
, and the joint velocity of the disturbance joint. The object is fixed on the table once the DR is finished. 


Through DR, we enhance the adaptability of the policy to variations in the action-force relationship. For example, instead of controlling the policy joint, the policy may generalize to control another joint of the manipulator, as this is a change to relative pose between the end-effector and the joint, which results in a variation in the action-force relationship. It may also generalize to objects with different stiffnesses, as those are also variations in action-force transition.

\subsection{Task-Balanced Network Updates (TBNU)}
\noindent
Although the contact locations are randomly sampled during training and have an equal probability of being on the two contact regions with mirrored action-force relationships, the mini-batches used to update the neural networks contain a small amount of data that may have a sampling imbalance. In experiments conducted without our balancing mechanism, we observed the learned policy with bias toward controlling one region instead of discovering a strategy to control both. 

Our approach for learning a balanced control policy is inspired by \emph{Model Agnostic Meta Learning (MAML)}~\cite{Finn2017maml} and class-balanced supervised learning~\cite{zhang2020class}. Specifically, we treat the control of each contact region as a different task and create two replay buffers instead of one. We sample a fixed length of transitions from each buffer and update the networks in sequence. Although it may not provide the few-shot retraining performance seen in MAML, we see TBNU as a stronger way than DR to keep our policies out of local minima that perform well for only one region.



\section{Experiments}
\subsection{Learning Performance}
\noindent
We trained the policy in simulation (MuJoCo \cite{todorov2014convex}) on a desktop computer with Ubuntu 18.04, an Intel i9-9900 CPU, and 32GB memory. A single agent was used to collect data for the policy. There were 400 training epochs, and in each epoch, we performed 100 policy rollouts each for 60 time-steps (2.4 million time-steps in total). Both actor (the policy) and critic networks have four hidden layers, with 64 neurons for the first three and 32 for the last layer. The activation function is ReLU \cite{agarap2018relu} for the hidden layers, and linear for the output layer. A joint speed limit is specified for the force controller to prevent simulation instability and hardware damage. Under this setup, training using PyTorch took about 4 hours (1.5 hours is enough for high scores).

For comparison purpose, we trained multiple policies with different design choices. All six policies listed below were trained using DDPG, HER, and DR, on top of which we varied the interaction history length, feature selection, and the use of TBNU:
\begin{itemize}
    \item \textbf{\textit{1obs}}: ``1obs'' means $N=1$ for the interaction history $\mathcal{H}$ in the observation. The policy observes all features $f_{n}$, $\Delta f_{n}$, and $\Delta^2 f_{n}$ and is trained with no TBNU. The same naming pattern applies to the rest of the policies. 
    \item \textbf{\textit{1obs\_noddf}}: $N=1$, trained with no TBNU, $\Delta^2 f_{n}$ is not observed.
    \item \textbf{\textit{1obs\_tbnu}}: $N=1$, TBNU is enabled.
    \item \textbf{\textit{3obs\_tbnu}}: $N=3$, TBNU is enabled.
    \item \textbf{\textit{3obs\_noddf\_tbnu}}: $N=3$, TBNU is enabled, $\Delta^2 f_{n}$ is not observed.
    \item \textbf{\textit{5obs\_tbnu}}: $N=5$, TBNU is enabled.
\end{itemize}

\noindent
Evaluation is done at the end of each training epoch. We perform 50 policy rollouts 
and compute the evaluation score. $\text{Score} = \overline{0.4 + \max(\text{reward}_{ij},-0.4)}$, where $i$ is the rollout number and $j$ is the time-step. Intuitively, the score reflects the step-wise reward with a force threshold, such that it is only counted when the policy brings the contact force within a $0.4N$ range of the target. From our experience, the $0.4N$ threshold offers a focused measuring window and strike a good balance between capturing the steady-state performance and maintaining the rising time information. 

\begin{figure}[]
    \vspace{0.2em}
    \includegraphics[width=0.485\textwidth]{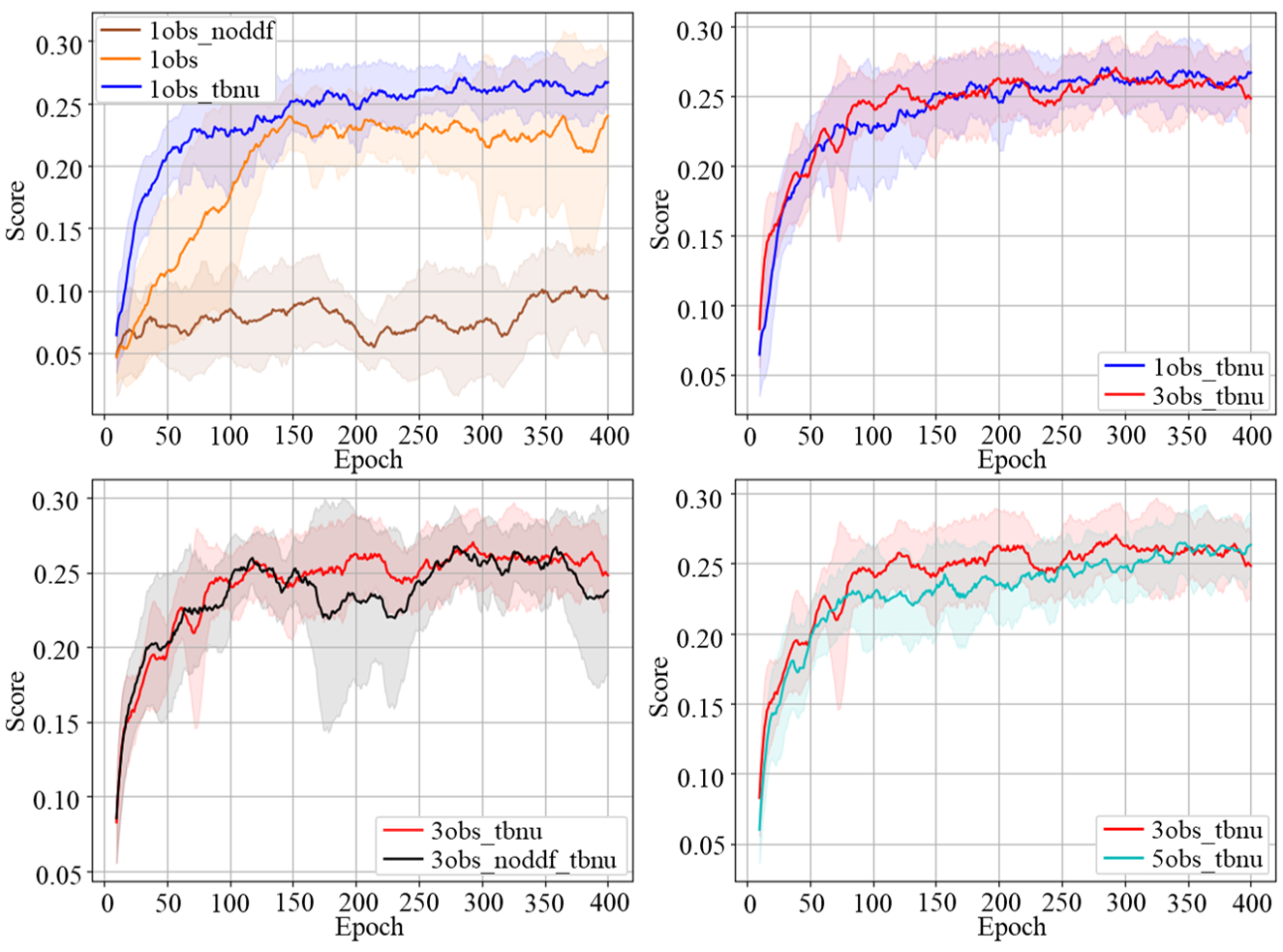}
    \caption{Learning curves (5 random seeds, mean and standard deviation are shown).}
    \label{fig:lc}
    \vspace{-1em}
\end{figure}

The plots in Fig.~\ref{fig:lc} shows the learning curves (10-epoch moving-average) for all six policies. In the upper left plot, policy \textit{1obs\_noddf} scored much lower than the other two, and \textit{1obs\_tbnu} learned faster than \textit{1obs} and scored higher. We checked the policies visually using the simulator, and found that \textit{1obs\_noddf} only worked for contacts in one region of the sensor, while the other two worked for both regions. This suggests that 1) the $\Delta^2 f_{n}$ observation is critical when $N=1$; and 2) TBNU helps in achieving force control for both contact regions. 

When $N$ is increased to 3, shown in the upper right plot, policy \textit{3obs\_tbnu} can achieve similarly high scores as \textit{1obs\_tbnu} but is faster and with less variations among random seeds. On the bottom left, when the $\Delta^2 f_{n}$ observation is removed, \textit{3obs\_noddf\_tbnu} can still achieve high scores but the training was much less stable. This suggests that the policy benefits from the direct observation of $\Delta^2 f_{n}$. Further increasing the observation history length may encourage the policy to overfit, which can be detrimental to the learning. For example, policy \textit{5obs\_tbnu} in the bottom right plot reaches high scores at a slower rate.

\begin{figure}[]
    \vspace{0.1em}
    \includegraphics[width=0.485\textwidth]{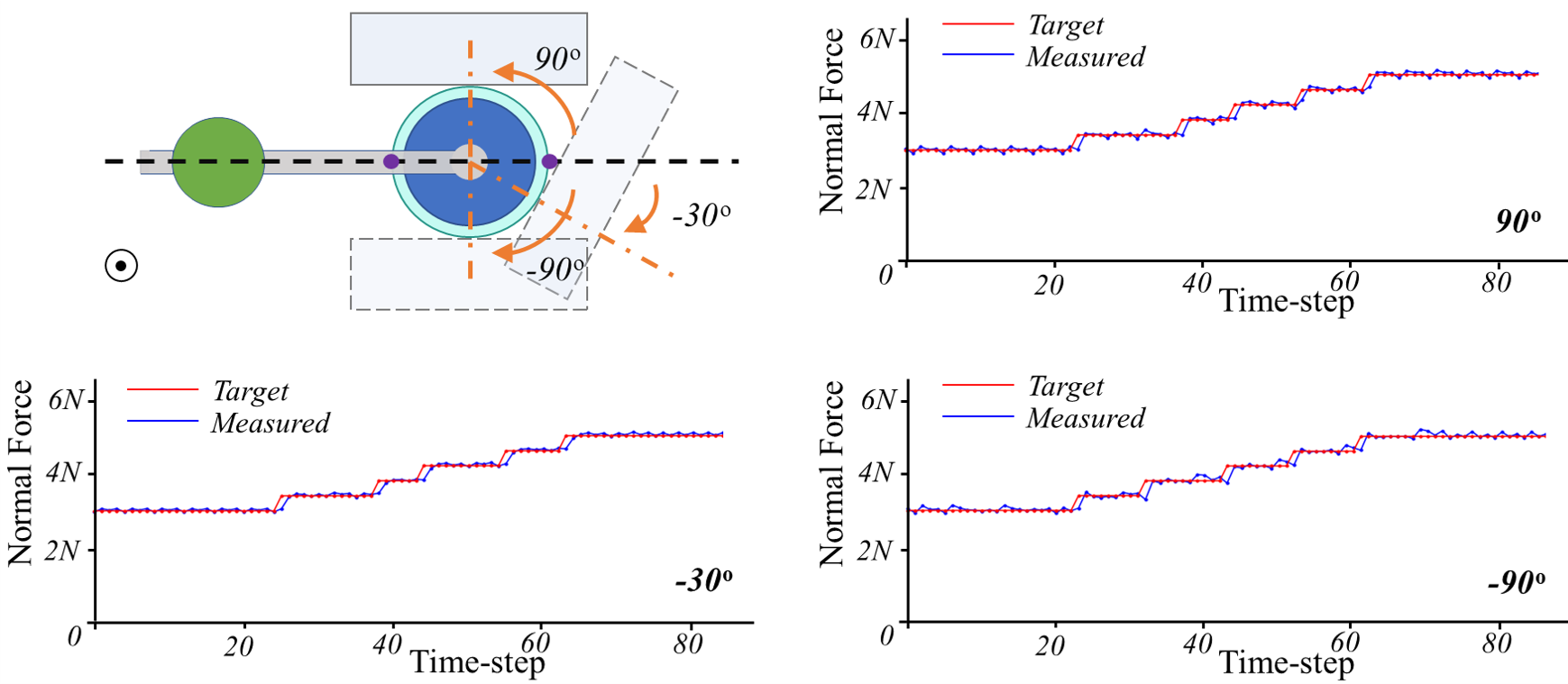}
    \caption{The policy can reach and maintain the normal contact force at various contact locations.}
    \label{fig:angle}
    \vspace{-1.5em}
\end{figure}

The evaluation scores demonstrate the quantitative performance of the learned policy in simulation.  
Fig.~\ref{fig:angle} shows examples of a high-score policy (\textit{3obs\_tbnu}, scored over 0.3) in action in simulation. The policy controls the green joint to reach and maintain the desired normal force at a single contact ($90^{o}$, $-90^{o}$, or $-30^{o}$). The initial contact force is set at $3N$, and we manually increase the target contact force, $0.4N$ per step, until it reaches $5N$. The plots show the target force in red color and the measured force in blue. The x-axis is the time-step, and the y-axis is the force (same apply to the rest of the plots in the paper). The policy was able to keep up with the target for all three contact angles, and there is no obvious performance difference in force-tracking. 

\subsection{Hardware Experiments}
We designed two real-world experiments to show how our force controller, in combination with robot motions, enables fine contact interactions. The hardware system (Fig.~\ref{fig:robot_GEN3}) consists of a Kinova GEN3 7-DOF robot arm and a cylindrical end-effector with a low-cost, flexible tactile sensor wrapped around its lateral surface. The sensor is a single-zone force-sensing resistor strip, not a tactile array, which only measures the magnitude of the normal force for a single contact. It cannot measure the location of the contact, the direction of the contact normal, or the friction force. We assume there is a single contact between the robot's end-effector and the object. The robot is posed close to the configuration shown in Fig.~\ref{fig:robot_GEN3} in the experiments, and the force control frequency is 5Hz due to actuation latency. Experimental results are reported below:

\begin{figure}[t]
    \centering
    \includegraphics[width=0.48\textwidth]{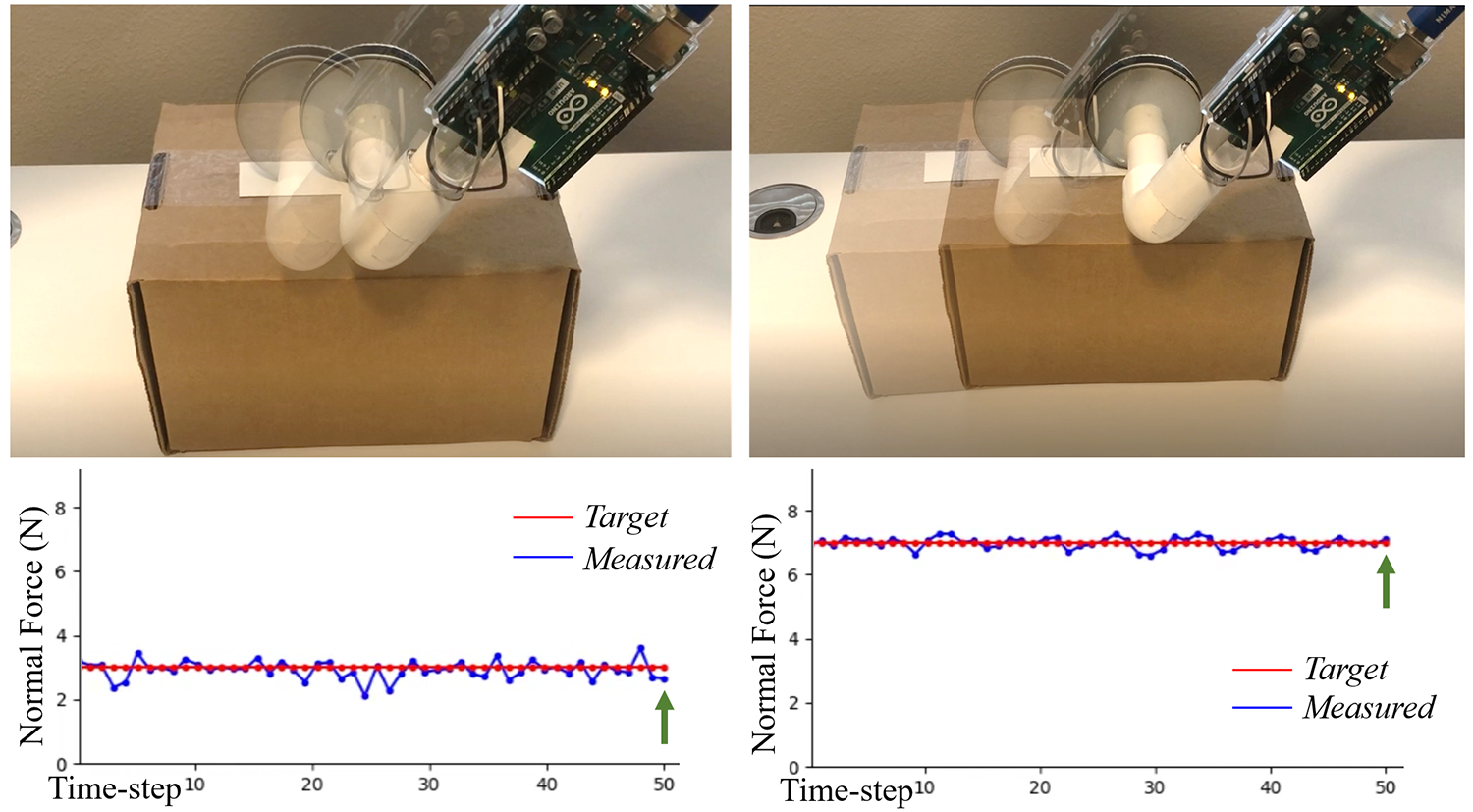}
    \caption{Left: Slipping contact, the end-effector slides to the left without moving the object while maintaining contact. Right: Sticking contact, the end-effector slides the box to the left. The plot below each photo shows the recorded force measurements, in which the green arrows indicate the current time-step.}
    \label{fig:slide_small}
\end{figure}

\subsubsection{Experiment 1 - sticking and slipping contacts}
As shown in Fig.~\ref{fig:slide_small}, initial contact is made between the end-effector and the cardboard box. The robot is commanded to move to the right by a velocity controller \cite{chen2018software}, facilitated by all seven joints. Our force controller is used to modify the joint velocity of the sixth joint to track the desired normal contact force $f_{n}^{d}$. When $f_{n}^{d}$ is low (Fig.~\ref{fig:slide_small}, left), the motion results in a slipping contact. The force controller ensures a robust slipping interaction, allowing the robot to explore the box's top surface. When $f_{n}^{d}$ is high (Fig.~\ref{fig:slide_small}, right), the slipping contact transitions into a sticking contact, and the box moves with the robot. More experiments with various motion directions and contour following of a curved surface can be found in the accompanying video.

\begin{figure}[h]
    \centering
    \includegraphics[width=0.48\textwidth]{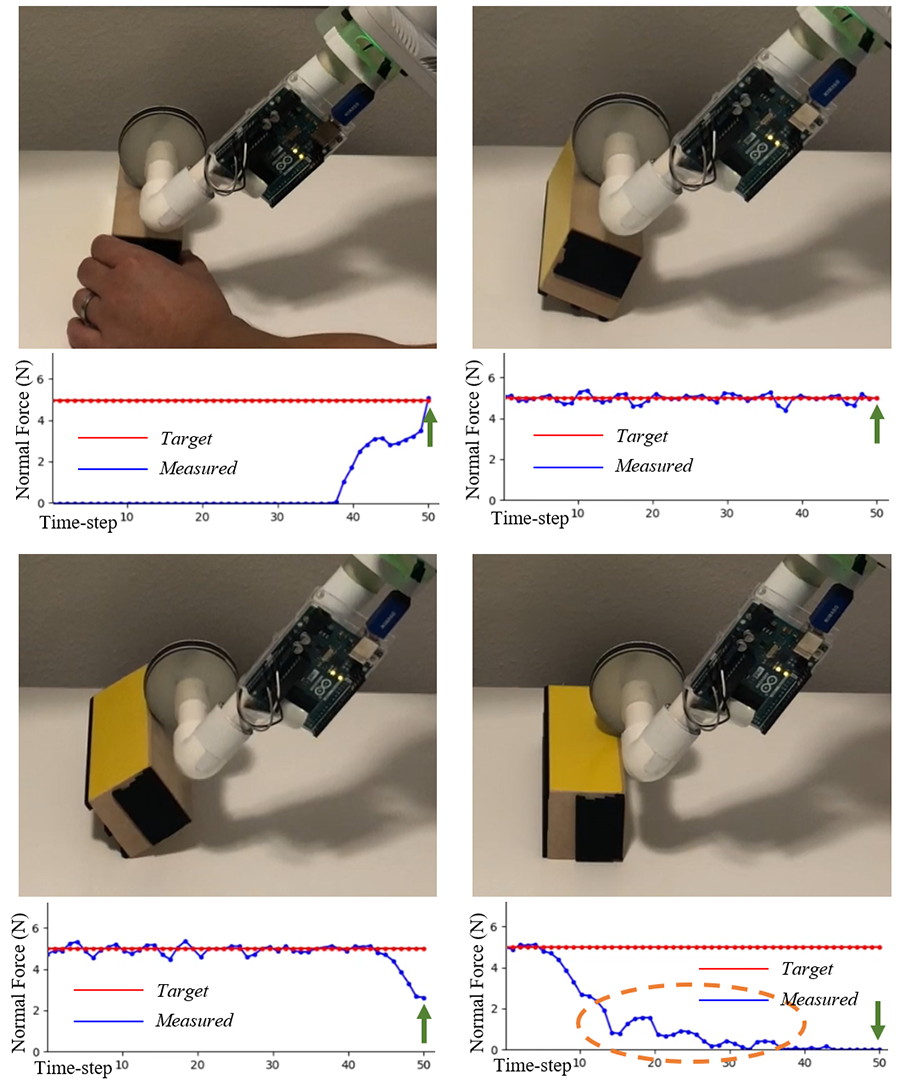}
    \caption{Tilting a block. From left to right, top to bottom: initial contact made, tilting started, just passed the mid-way, fully tilted.}
    \label{fig:tilt}
    \vspace{-1em}
\end{figure}

\subsubsection{Experiment 2 - rolling contact}
The task in experiment 1 has orthogonal motion and force control directions, such that the parallel motion controller causes minimal interference to the force controller. In this experiment, we use a block tilting task (shown in Fig.~\ref{fig:tilt}) to demonstrate our force controller's robustness to disturbances in the force control direction when the contact is rolling.

In the experiment, the robot's fourth joint tilts up at 0.24 degrees/s, which approximately moves the end-effector to the right. At the contact, this movement disturbs the normal deformation (recall the lumped spring at the contact point), and the disturbance level changes as the object tilts and the contact normal direction rotates. The control policy is deployed on the sixth joint, and it needs to resist disturbance and maintain the desired normal contact force. After the initial contact, the policy achieves the desired normal force, which results in a sticking contact with the wooden block. As the fourth joint rotates, the end-effector moves to the right, and the block tilt because of the sticking contact. During tilting, the sticking contact transforms into a rolling contact. After the block tilts past the self-balancing configuration, the normal contact force decreases as gravity becomes the major source of the force. The policy then tries to increase the force by pushing back (highlighted by the orange circle), but the block slips away as the normal force continues to decrease and finally tilts completely.

We report baseline comparisons, additional experiments, and implementation considerations in \cite{cui2023icrasup}.

\section{Conclusion and Future Work}

We developed a low-cost solution to the marginally-informed normal force control problem in this study. Our learning results demonstrated the validity of key learning design choices, and our hardware experiments showed that the learned controller could be implemented in parallel with a motion controller to be deployed on a real-world robotic system to facilitate fine contact interactions. Although being the first of its kind, the proposed solution is far from perfect, which encourages further study:

\begin{itemize}
    \item Although already useful, our learned policy only controls a single robot joint. The force control capacity depends on the contact location, \textit{i.e.}, the policy loses control over the normal force when boundary singularity occurs. It is less likely for this to occur if the policy can control all robot joints simultaneously.
    \item This study focused on force control of a single contact point using a robot arm as a ``dexterous finger.'' A future direction would be using multiple dexterous fingers to enable in-hand dexterous manipulation.
    \item The learned force controller, combined with a parallel motion controller, can be used as a building block for learning higher-level manipulation skills requiring fine contact interactions. For example, the contact-maintaining nature of our controller may enable more efficient exploration compared to intermittent contacts in reinforcement learning. The ways our controller can be used in learning and planning for manipulation tasks is a rich field for future research. 
\end{itemize}

\bibliographystyle{ieeetr}
\bibliography{references.bib}

\end{document}